\documentclass[letterpaper]{article} 
\usepackage{aaai2027}  
\usepackage{times}  
\usepackage{helvet}  
\usepackage{courier}  
\usepackage[hyphens]{url}  
\usepackage{graphicx} 
\usepackage{natbib}  
\usepackage{caption} 
\usepackage{booktabs}
\usepackage{amsmath}
\usepackage{amssymb}
\title{DualMem: Bypassing the Objectness Bottleneck for Calibrated Unknown-Stream Filtering in Open-World Object Detection}
\author{
    Yingjun Xiao\textsuperscript{\rm 1},
    Xi Chen\textsuperscript{\rm 2},
    Gang Fang\textsuperscript{\rm 3}\thanks{Corresponding author.},
    Siyuan Chen\textsuperscript{\rm 2}
}
\affiliations{
    \textsuperscript{\rm 1}School of Artificial Intelligence, Guangzhou University, Guangzhou, China\\
    \textsuperscript{\rm 2}School of Computer Science and Cyber Engineering, Guangzhou University, Guangzhou, China\\
    \textsuperscript{\rm 3}Institute of Computing Science and Technology, Guangzhou University, Guangzhou, China
}

\newcommand{\methodname}{DualMem}

\begin{document}

\maketitle

\begin{abstract}
Open-world object detection (OWOD) requires detectors to localize known classes
and flag unknown objects for future incremental learning. We observe that the
unknown prediction streams of strong OWOD detectors are heavily polluted:
across PROB, OW-DETR, and HypOW on M-OWODB, future-task positive unknowns
account for less than 10\% of unknown predictions, while background false
positives occupy 46--71\%. We show this is not a missing-information problem
but an \emph{information bottleneck} at the objectness head: on PROB Task~1, a
linear probe on the 256-D decoder query reaches $\mathrm{AUROC}=0.908$ for
positive-versus-negative unknown discrimination, while the final
one-dimensional objectness scalar drops to $0.642$. A frozen SigLIP feature, with no access to the detector, independently restores much of this proposal-level separability at the filtering stage
($\mathrm{AUROC}=0.871$). Motivated by this, we propose \methodname{}, a calibrated post-hoc filter
that assumes a small image-disjoint annotated calibration split for
held-out future-task objects and performs a \emph{non-parametric likelihood
ratio test} in the frozen SigLIP feature space. It uses a $k$-nearest-neighbor positive memory to guard future-task objects and a negative memory to suppress background-like proposals. The decision threshold is selected by Neyman--Pearson calibration, exposing an explicit, user-controllable
trade-off between false-unknown suppression and novel recall. Across PROB, OW-DETR, and HypOW on M-OWODB Task~1, \methodname{} reduces
background-type false unknown proposals per image by
$44.9\%$--$66.3\%$ (mean $56.6\%$). On PROB Task~1, it more than doubles
the reduction achieved by a natural K-means prototype baseline, while leaving known-class mAP unchanged because known detections bypass the
filter.
\end{abstract}

\section{Introduction}
\label{sec:introduction}

Open-world object detection (OWOD) asks a detector to operate beyond a
fixed label set: it must localize known classes, flag unknown objects, and
later incorporate these unknowns through incremental learning. Recent OWOD
detectors have made steady progress on unknown recall, from early
open-world formulations to DETR-style, probabilistic, hyperbolic, and
prototype/cascade variants~\cite{ore,detr,ow-detr,prob,hypow,uc_owod,
cat_owod,owod_prototype}. Yet a closer look at what these detectors
actually output reveals a structural problem: the unknown prediction stream
is heavily polluted by detections that are not future-task objects.

\begin{figure}[t]
    \centering
    \includegraphics[width=0.95\linewidth]{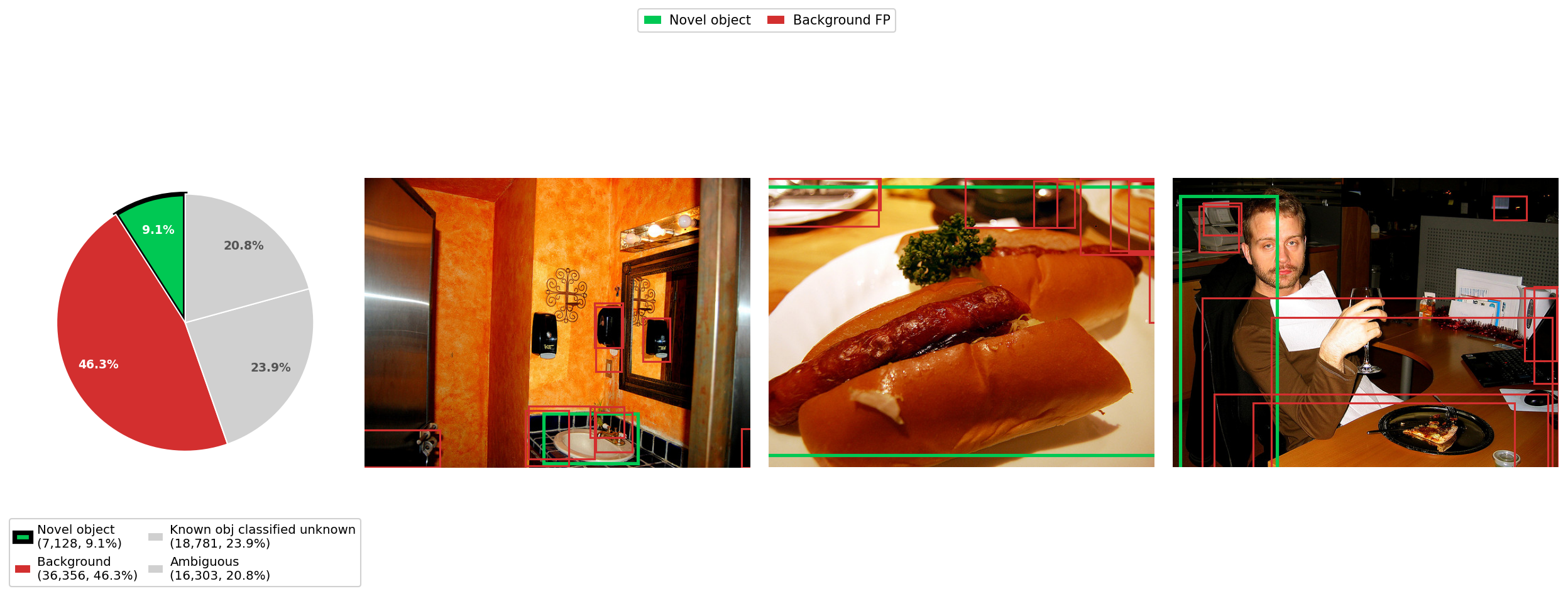}
    \caption{
    Motivation. The unknown prediction stream of a strong OWOD detector is
    dominated by non-novel outputs. On PROB Task~1, only $9.1\%$ of unknown
    predictions are positive unknowns, while most predictions are background
    false positives or known objects predicted as unknown.
    }
    \label{fig:motivation}
\end{figure}

We decompose the unknown stream into four categories: positive unknowns,
which match future-task ground-truth objects; negative unknowns, which do
not match any ground truth; known-as-unknown predictions, which overlap
current known objects; and ambiguous predictions. Figure~\ref{fig:motivation}
illustrates this decomposition for PROB on M-OWODB Task~1. Although PROB is
a strong OWOD detector, only $9.1\%$ of its unknown predictions on the
val2017 split are positive unknowns, while $46.3\%$ are negative unknowns
and $23.9\%$ are known objects mislabeled as unknown. This pattern is not
specific to PROB: across PROB, OW-DETR, and HypOW, positive unknowns account
for less than $10\%$ of the unknown stream in all nine detector-task cells
from Task~1 to Task~3. Thus, the practical challenge is not only to recall
unknown objects, but also to clean a prediction stream dominated by
non-novel outputs.

Why do detector outputs fail to separate positive and negative unknowns? A
natural explanation is that the detector may simply lack the information
needed for this distinction. Our analysis shows the opposite. On PROB
Task~1, the detector's final objectness score achieves only
$\mathrm{AUROC}=0.642$ for positive-vs-negative unknown discrimination,
indicating weak separability at the output. However, a linear probe trained
on the detector's $256$-D decoder query features reaches
$\mathrm{AUROC}=0.908 \pm 0.009$ under five-fold cross-validation. The
information needed to distinguish positive from negative unknowns exists
inside the detector, but it is largely lost when the decoder query is
compressed into a one-dimensional objectness scalar. A frozen SigLIP
ViT-B/16 crop feature, with no access to the detector's internal state,
achieves $\mathrm{AUROC}=0.871 \pm 0.013$, recovering much of the lost
discriminability from an external representation. These results identify
the objectness head as an information bottleneck: the detector contains
useful evidence, but its final unknown confidence signal discards it.

This observation motivates \methodname{}, a post-hoc unknown-stream filter
that treats the OWOD detector as fixed and applies a non-parametric
likelihood ratio test in the frozen SigLIP feature space. Two
memories---a positive memory of future-task proposals and a negative memory
of background false positives---define class-conditional kernel density
estimators built directly on calibration features ($k$-nearest neighbor; no
centroid quantization). The decision threshold is selected by Neyman--Pearson-style calibration to
target a novel mis-suppression budget $\alpha$, turning an opaque magic
number into an explicit precision--recall trade-off. Across three architecturally distinct
detectors---PROB, OW-DETR, and HypOW---\methodname{} reduces FUPI by
$56.6\%$ on average (range $44.9$--$66.3\%$), with U-Recall preserved within $3.4$ percentage points and known-class mAP unchanged because known detections bypass the filter. We deliberately choose the minimal post-hoc
construction---no detector retraining, no new losses---so that 
any improvement can be attributed to bypassing the scalar objectness
bottleneck and restoring proposal-level separability at the filtering stage.

Our contributions are threefold:

\begin{itemize}
    \item \textbf{Diagnosis: scalar objectness is a cross-detector
    bottleneck.} Across PROB, OW-DETR, and HypOW on M-OWODB Task 1, the
    final objectness scalar is a weak positive-vs-negative unknown
    discriminator, with AUROC $0.642$, $0.526$, and $0.516$, respectively,
    and high positive/negative overlap. In contrast, detector-internal
    proposal features achieve AUROC $0.908$, $0.855$, and $0.881$, while
    frozen SigLIP crop features achieve AUROC $0.871$, $0.828$, and
    $0.888$. These results show that the unknown stream is not merely
    information-poor; rather, discriminative proposal-level information is
    not exposed by the final one-dimensional objectness interface.

    \item \textbf{Method: \methodname{} as a minimal post-hoc bypass of the objectness bottleneck.} We treat unknown-stream filtering as a non-parametric
    likelihood ratio test in a frozen SigLIP feature space. Two
    $k$-nearest-neighbor memories---one for positive unknowns, one for
    background---define class-conditional density estimators; the suppression
    threshold is selected by Neyman--Pearson calibration to target the
    false-suppression rate at a user-specified $\alpha$. On M-OWODB Task~1 with $\alpha=0.10$, this reduces FUPI by $58.6\%$
    on PROB, $44.9\%$ on OW-DETR, and $66.3\%$ on HypOW. On PROB Task~1,
    it more than doubles the reduction achieved by a K-means prototype baseline
    (\S\ref{sec:dual_memory_ablation}).

    \item \textbf{Evaluation: a three-axis framework for post-hoc
    unknown-stream filters.} Since \methodname{} leaves the known-class
    stream untouched, classical WI and A-OSE are structurally insensitive
    to its main operation. We adopt three complementary
    axes---FUPI (background-type false-unknown density), UDP (unknown detection
    precision)~\cite{re_owod}, and NMH (novel mis-suppression
    harm)---that together distinguish healthy stream cleanup from
    aggressive deletion hidden by recall alone.
\end{itemize}

\section{Related Work}
\label{sec:related_work}

\paragraph{Open-world object detection.}
OWOD extends detection beyond a fixed label space, building on open-set
recognition and open-set object detection but adding explicit unknown
localization and incremental class learning~\cite{open_set_recognition,
open_set_detection,ore}. Methods improve unknown discovery through energy
scores~\cite{ore}, attention-driven pseudo-labels~\cite{ow-detr},
probabilistic objectness~\cite{prob}, hyperbolic embeddings~\cite{hypow},
unknown-class supervision~\cite{uc_owod}, localization-identification
cascades~\cite{cat_owod}, driving-specific reformulations~\cite{owod_autonomous},
prototype learning~\cite{owod_prototype}, or training-time objectives on
proposal generation~\cite{unsniffer,randbox}. All of these determine unknown
predictions from \emph{detector-internal} signals. \methodname{} is
orthogonal: it filters the unknown stream of a \emph{frozen} detector with an
external critic, requiring no retraining and applicable to proprietary
checkpoints.

\paragraph{Open-vocabulary detection and vision-language features.}
Open-vocabulary detectors use language supervision to recognize categories
outside a closed training vocabulary~\cite{vild,owlvit,yoloworld}. This
setting is related but distinct from OWOD: the test-time category names are
available, whereas OWOD must flag unknown objects before their labels are
introduced. \methodname{} uses vision-language pretraining only as a frozen
proposal critic, not as a classifier over text prompts.

\paragraph{Post-hoc OOD detection.}
On image classification, post-hoc OOD methods have shown that frozen
representations and output scores carry strong OOD signal without modifying
the classifier, from maximum-softmax and temperature-scaled baselines to
Mahalanobis, energy, and nearest-neighbor scores~\cite{hendrycks_ood,odin,
mahalanobis_ood,energy_ood,knn_ood,nnguide}. We adapt this idea to the
proposal level under the OWOD-specific constraint that suppressing
OOD-looking predictions must not erase future-task objects. The
$k$-nearest-neighbor construction of our memories is most closely related
to KNN-based OOD~\cite{knn_ood}, but the dual-memory rule with a positive
guard is, to our knowledge, new in this setting.

\paragraph{OWOD evaluation.}
Traditional metrics---U-Recall, Wilderness Impact, A-OSE~%
\cite{ore,ow-detr,prob}---measure either novel recall or contamination of
the known stream. RE-OWOD~\cite{re_owod} introduces UDP for unknown
precision. We complement UDP with FUPI (absolute background-type false-unknown density)
and NMH (false-suppression cost), which together describe the operating
space of post-hoc filters operating only on the unknown stream.

\section{Why Detector-Internal Filtering Fails}
\label{sec:why_internal_filtering_fails}

\subsection{The Pollution Problem}
\label{sec:pollution_problem}

We first examine what an OWOD detector actually emits in its unknown
prediction stream. For each unknown prediction, we assign it to one of four
groups by priority. A \emph{positive unknown} is first defined as a prediction
with $\mathrm{IoU}\geq 0.5$ to a future-task ground-truth object. Among the
remaining predictions, a \emph{known-as-unknown} prediction overlaps a
current-known ground-truth object with $\mathrm{IoU}\geq 0.5$. A
\emph{negative unknown} has maximum IoU below $0.3$ with any ground-truth
object, and therefore corresponds to background. The remaining predictions
are treated as \emph{ambiguous}, covering intermediate-overlap cases.

Table~\ref{tab:unknown_decomposition} shows this decomposition across three
OWOD detectors and four M-OWODB tasks. The main observation is that positive unknowns form only a small
minority of the unknown stream. Across PROB, OW-DETR, and HypOW, all nine
detector-task cells from Task~1 to Task~3 contain fewer than $10\%$
positive unknowns. Task~4 contains no future-task objects by construction,
so its positive-unknown rate is $0.0\%$. In contrast, negative unknowns
alone account for $46$--$71\%$ of the stream. This pollution appears across
architecturally distinct detectors, suggesting a structural property of
detector-internal unknown filtering rather than an idiosyncratic failure of
one method.

\begin{table}[t]
    \centering
    \small
    \setlength{\tabcolsep}{4pt}
        \resizebox{\linewidth}{!}{
    \begin{tabular}{llrrrr}
        \toprule
        Detector & Task & Positive U. & Negative U. & Known-as-U. & Ambiguous \\
        \midrule
        PROB    & T1 & 9.1  & 46.3 & 23.9 & 20.8 \\
        PROB    & T2 & 7.8  & 51.9 & 20.6 & 19.6 \\
        PROB    & T3 & 2.0  & 52.7 & 27.3 & 17.9 \\
        PROB    & T4 & 0.0  & 49.1 & 33.3 & 17.6 \\
        \midrule
        OW-DETR & T1 & 2.1  & 66.9 & 18.7 & 12.3 \\
        OW-DETR & T2 & 2.5  & 63.7 & 22.0 & 11.8 \\
        OW-DETR & T3 & 2.4  & 60.3 & 25.7 & 11.6 \\
        OW-DETR & T4 & 0.0  & 70.6 & 22.4 & 6.9  \\
        \midrule
        HypOW   & T1 & 9.6  & 46.2 & 22.5 & 21.6 \\
        HypOW   & T2 & 6.8  & 56.6 & 17.4 & 19.1 \\
        HypOW   & T3 & 1.7  & 61.2 & 19.2 & 17.9 \\
        HypOW   & T4 & 0.0  & 56.0 & 25.7 & 18.3 \\
        \bottomrule
    \end{tabular}
    }
\caption{
    Unknown-stream decomposition on M-OWODB. Positive unknowns correspond
    to future-task objects, negative unknowns correspond to background
    false positives, known-as-unknown predictions overlap current-known
    objects, and ambiguous predictions fall outside these clean cases.
    }
    \label{tab:unknown_decomposition}
\end{table}

\subsection{The Information Bottleneck}
\label{sec:information_bottleneck}

The polluted unknown stream raises a natural question: do OWOD
detectors lack the information needed to distinguish positive unknowns
from negative unknowns, or is this information present in richer
proposal representations but discarded by the final objectness output?
We answer this question with a cross-detector binary probing experiment
on M-OWODB Task 1. For each detector, we label raw unknown proposals as
positive if they match future-task ground truth and negative if they
are background false unknowns; known-as-unknown and ambiguous proposals
are excluded. We compare three signals: the detector's final
one-dimensional objectness score, a detector-internal proposal feature,
and a frozen SigLIP crop feature. Feature representations are evaluated
with logistic-regression linear probes under five-fold image-level
GroupKFold.

Table~\ref{tab:cross_detector_probe} shows a consistent pattern across
PROB, OW-DETR, and HypOW. The final objectness scalar is weak:
objectness AUROC is $0.642$ on PROB, $0.526$ on OW-DETR, and $0.516$ on
HypOW, with high positive/negative overlap. In contrast,
detector-internal proposal features are strongly discriminative,
reaching AUROC $0.908$, $0.855$, and $0.881$, respectively. Frozen
SigLIP crop features also retain substantial separability, reaching
AUROC $0.871$, $0.828$, and $0.888$.

These results support a cross-detector objectness-bottleneck diagnosis.
The unknown stream is not simply devoid of information: richer
proposal-level representations, whether internal or external, separate
positive and negative unknowns far better than the final scalar
objectness output. However, high supervised linear separability does
not by itself guarantee that a feature is a good non-parametric memory
critic. Section~\ref{sec:critic_ablation} shows that, on PROB, the
detector query has the highest linear AUROC but is inferior to SigLIP
under raw cosine kNN filtering. This motivates DualMem's use of a
frozen external critic whose feature geometry is better aligned with
nearest-neighbor retrieval.

\begin{table*}[t]
\centering
\caption{
Cross-detector positive-vs-negative unknown separability on
M-OWODB Task 1. The binary labels are
$\mathrm{positive\_unknown}=1$ and $\mathrm{negative\_unknown}=0$;
known-as-unknown and ambiguous proposals are excluded. AUROC for SigLIP
and internal features is measured with five-fold image-level GroupKFold
linear probing. OVL is the overlap coefficient of the corresponding
one-dimensional score distributions: raw objectness for the objectness
column and linear-probe logits for SigLIP and internal features.
}
\label{tab:cross_detector_probe}
\resizebox{\textwidth}{!}{
\begin{tabular}{lrrrrrrrr}
\toprule
Detector & Pos. & Neg. & Obj. AUROC & Obj. OVL
& SigLIP AUROC & SigLIP OVL
& Internal AUROC & Internal OVL \\
\midrule
PROB
& 7,128 & 36,356 & 0.642 & 0.771
& $0.871 \pm 0.013$ & 0.418
& $0.908 \pm 0.009$ & 0.348 \\
OW-DETR
& 362 & 11,757 & 0.526 & 0.913
& $0.828 \pm 0.031$ & 0.262
& $0.855 \pm 0.028$ & 0.292 \\
HypOW
& 4,469 & 21,412 & 0.516 & 0.952
& $0.888 \pm 0.005$ & 0.372
& $0.881 \pm 0.007$ & 0.397 \\
\bottomrule
\end{tabular}}
\end{table*}

\begin{figure}[t]
    \centering
    \includegraphics[width=\linewidth]{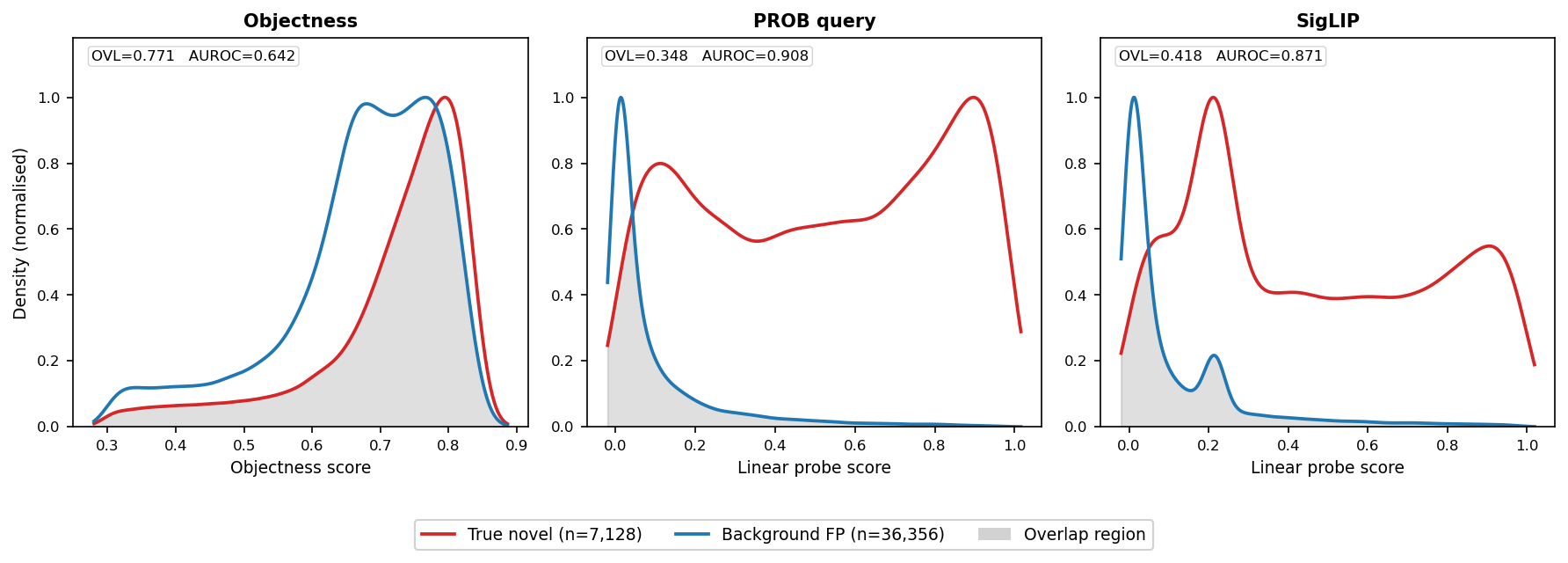}
    \caption{
    PROB Task 1 feature separability for positive and negative unknown
    proposals. The objectness score exhibits heavy distribution overlap,
    whereas the decoder query and frozen SigLIP crop feature provide much
    stronger separation under a linear probe. Cross-detector quantitative
    results are reported in Table~\ref{tab:cross_detector_probe}.
    }
    \label{fig:linear_probe_kde}
\end{figure}

This bottleneck is not unique to PROB: the objectness-score overlap coefficient between positive and negative unknowns is 0.771 for PROB, 0.913 for OW-DETR, and 0.952 for HypOW. Detector-output confidence is
a weak signal across architectures. The pollution and linear-probe results
jointly suggest that the failure is not missing information but
\emph{discarded} information at the objectness interface, motivating a
post-hoc filter that bypasses this scalar entirely and uses frozen external
features as an independent critic.

\section{\methodname{} Method}
\label{sec:dualmem_method}

\begin{figure*}[t]
    \centering
    \includegraphics[width=0.98\textwidth]{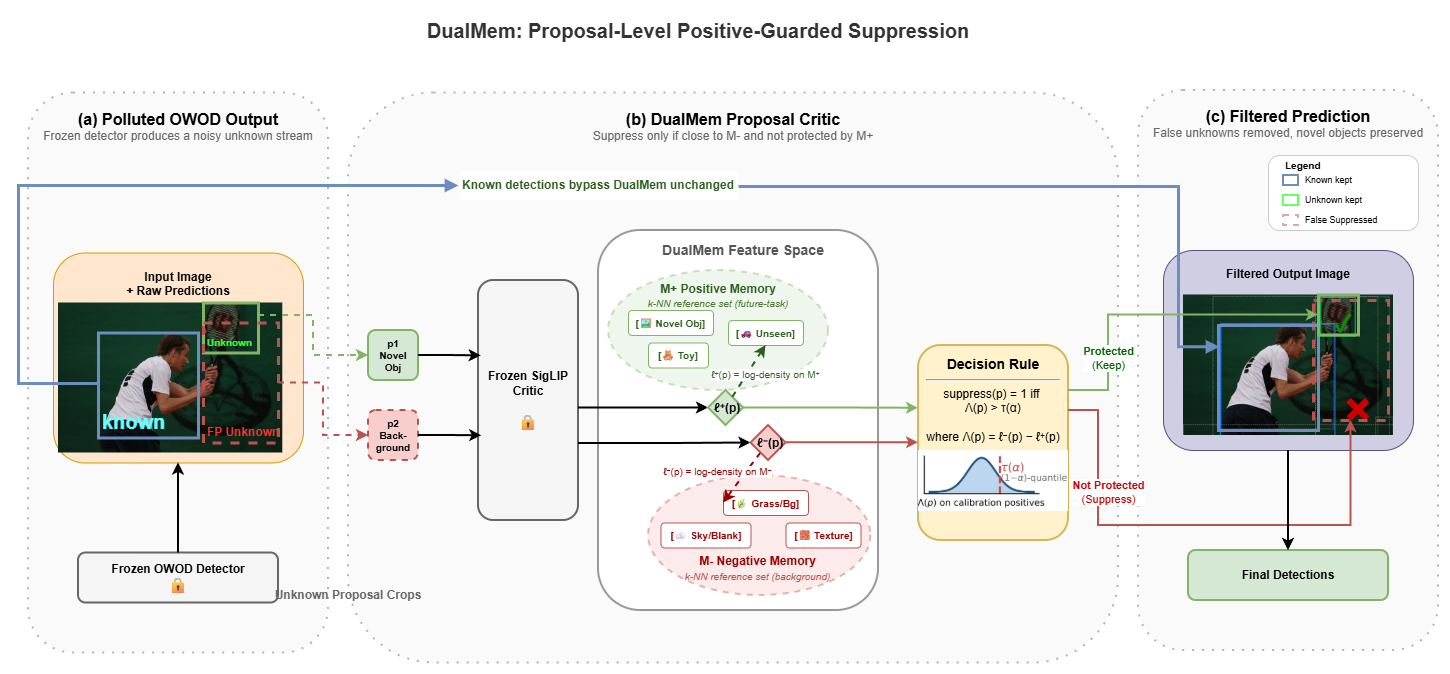}
\caption{
Overview of \methodname{}. A frozen OWOD detector emits an unknown stream.
\methodname{} crops each unknown proposal, embeds it with a frozen SigLIP
critic, compares it against positive and negative k-NN memories, and applies
an NP-calibrated likelihood-ratio threshold $\tau(\alpha)$. Known detections
bypass the filter entirely.
}
    \label{fig:dualmem_overview}
\end{figure*}

\methodname{} is a post-hoc filter that bypasses the scalar objectness
bottleneck by applying a non-parametric likelihood ratio test in a frozen
SigLIP feature space. We describe its three
components: a $k$-nearest-neighbor dual memory
(\S\ref{sec:method_memory}), a likelihood ratio decision rule
(\S\ref{sec:method_lrt}), and Neyman--Pearson threshold calibration
(\S\ref{sec:method_np}).

\subsection{Dual $k$-NN Memory in the SigLIP Feature Space}
\label{sec:method_memory}

\paragraph{Calibration assumption.}
DualMem assumes access to a small image-disjoint calibration split
with box-level annotations for held-out future-task objects. These
annotations are used only to construct and calibrate the post-hoc
filter, and are never used to train or fine-tune the OWOD detector.
Thus, our setting should be interpreted as calibrated post-hoc
deployment rather than fully annotation-free unknown discovery. In
deployment scenarios without annotated novel calibration boxes, one
would need to replace $M^+$ with mined high-confidence object proposals
or with a proxy positive memory; we leave this annotation-free variant
to future work.

Given a frozen OWOD detector and an image-disjoint calibration split, we
collect the detector's unknown proposals and assign each to one of the four
categories defined in Section~\ref{sec:pollution_problem}. We further split
the calibration positives into a memory subset
$\mathcal{P}^+_{\mathrm{mem}}$ and a threshold-calibration subset
$\mathcal{P}^+_{\mathrm{thr}}$. The positive memory is built only from
$\mathcal{P}^+_{\mathrm{mem}}$, while $\mathcal{P}^+_{\mathrm{thr}}$ is
reserved for choosing $\tau(\alpha)$. The negative memory
$\mathcal{P}^-_{\mathrm{mem}}$ contains background false positives from the
calibration split. For each proposal $p$, we crop the corresponding image
region and extract an $L_2$-normalized frozen SigLIP ViT-B/16
feature~\cite{vit,siglip}, $f(p)\in\mathbb{R}^{768}$.

We define two reference memories directly from these features, with no
centroid quantization:
\begin{equation}
    \mathcal{M}^+ = \{f(p) : p \in \mathcal{P}^+_{\mathrm{mem}}\},
    \quad
    \mathcal{M}^- = \{f(p) : p \in \mathcal{P}^-_{\mathrm{mem}}\}.
\end{equation}
This $k$-nearest-neighbor construction differs from prototype-based filters
(e.g., $K$-means centroids): every calibration proposal participates in
the density estimate, eliminating compression loss at the memory side and
following classical $k$-NN non-parametric density estimation and its modern
use in OOD detection~\cite{knn_density,knn_ood}. In our 
experiments, $|\mathcal{M}^-| \approx 30{,}000$ and 
$|\mathcal{M}^+|$ ranges from $\sim$$0.7$K (OW-DETR) to 
$\sim8$K (PROB) across detectors. We use frozen SigLIP 
features as the critic; a natural alternative is the detector's 
own decoder query, which carries strong supervised separability 
($\mathrm{AUROC}=0.908$, Table~\ref{tab:cross_detector_probe}) but, as 
we show in \S\ref{sec:critic_ablation}, proves a markedly weaker 
critic under raw cosine retrieval. We return to this 
empirical decoupling between supervised AUROC and memory-critic 
effectiveness in the critic-choice ablation.

\subsection{Non-Parametric Likelihood Ratio Test}
\label{sec:method_lrt}

For a test proposal $p$, we estimate the class-conditional log-density of
$f(p)$ under each memory by a temperature-scaled $k$-NN kernel estimator:
\begin{equation}
    \hat{\ell}^c(p) \;=\; \log \!\sum_{i=1}^{k} \exp\!\left(
    \frac{\cos\!\left(f(p),\, f(p_i^{c})\right)}{T}
    \right) - \log k,
    \label{eq:kde_score}
\end{equation}
where $c \in \{+, -\}$, $p_i^{c}$ are the top-$k$ nearest neighbors of
$f(p)$ in $\mathcal{M}^{c}$ under cosine similarity, and $T$ is a
temperature controlling kernel smoothness. The decision rule is a
single-threshold likelihood ratio test:
\begin{equation}
\begin{array}{rcl}
    \Lambda(p) &=& \hat{\ell}^{-}(p) - \hat{\ell}^{+}(p),\\
    \mathrm{suppress}(p) &=& \mathbf{1}\!\left[\Lambda(p) > \tau\right].
\end{array}
\label{eq:lrt_rule}
\end{equation}
This formulation has two convenient properties. First, as $T\to 0$, the
logsumexp in Eq.~\ref{eq:kde_score} reduces to a maximum and $\Lambda(p)$
collapses to a difference of single-prototype cosine similarities; the
rule becomes equivalent to a max-cosine positive-guarded suppression.
Larger $T$ smooths the kernel and aggregates evidence over multiple
neighbors, recovering otherwise brittle decisions on sparse memory
regions. Second, the same rule exposes a single, monotonic decision
variable $\Lambda(p)$, which makes principled threshold selection
possible (\S\ref{sec:method_np}). We use $T=0.05$ and $k=25$ as defaults.
Moderate changes to T and k did not alter the qualitative ranking of methods.

\subsection{Neyman--Pearson Threshold Calibration}
\label{sec:method_np}

A naive choice of $\tau$ is a magic number. We instead determine $\tau$
on the calibration split using the Neyman--Pearson criterion~\cite{neyman_pearson}:
given a
user-specified false-suppression budget $\alpha$ on positive unknowns,
choose the smallest $\tau$ such that
\begin{equation}
\Pr_{p\in \mathcal{P}^+_{\mathrm{thr}}}[\Lambda(p)>\tau]\leq \alpha,
\end{equation}
where $\mathcal{P}^+_{\mathrm{thr}}$ is a hold-out portion of the calibration
positives, disjoint from the memories. Operationally,
$\tau(\alpha)$ is the $(1{-}\alpha)$-quantile of
$\{\Lambda(p) : p \in \mathcal{P}^+_{\mathrm{thr}}\}$. This NP-style calibration selects the most aggressive threshold on the
calibration positives under a user-specified false-suppression budget. When
the k-NN score approximates a likelihood ratio and the calibration and test
distributions match, it follows the Neyman--Pearson operating principle;
under distribution shift, the realized test-time NMH may deviate from $\alpha$.

We use $\alpha=0.10$ throughout the main results. The $\alpha$-sweep
in Section~\ref{sec:alpha_sweep} shows that the user can trade FUPI for
U-Recall in a predictable, monotonic manner. Because the calibration
distribution can differ from the test distribution, actual test-time NMH
may deviate from $\alpha$; in practice we observe NMH $\le \alpha$ on
PROB and OW-DETR but a positive gap on HypOW (\S\ref{sec:main_results}),
which can be mitigated by per-detector validation and, if necessary, a more
conservative choice of $\alpha$.

\section{Experiments}
\label{sec:experiments}

\paragraph{Setup.}
We evaluate \methodname{} on M-OWODB across three architecturally distinct
    OWOD detectors: PROB~\cite{prob}, OW-DETR~\cite{ow-detr}, and
HypOW~\cite{hypow}. Calibration data come from a 20\% random subset of
COCO train2017 (image-disjoint from val2017); test evaluation is on COCO
val2017 following standard OWOD protocol. Unless noted, we use SigLIP
ViT-B/16~\cite{vit,siglip} as the frozen critic, $T=0.05$, $k=25$, and
$\alpha=0.10$. All
metrics are reported under a unified protocol that scores
unknown predictions against future-task ground truth.

\paragraph{A note on classical OWOD metrics.}
Because \methodname{} operates only on the unknown stream and leaves known
predictions unchanged, the classical WI and A-OSE, which measure unknown
contamination in the known-class stream, are structurally insensitive to
the main operation of our method. We therefore primarily report FUPI, UDP, NMH, and U-Recall, which directly characterize unknown-stream filtering.

\paragraph{Evaluation labels and metrics.}
For each detector-emitted unknown prediction $d$, we compute its IoU
with current-task known ground truth $G_{\mathrm{k}}$ and future-task
unknown ground truth $G_{\mathrm{u}}$. We assign proposal labels by the
following priority:
\[
\ell(d)=
\begin{cases}
\mathrm{pos}, &
\max_{g\in G_u}\mathrm{IoU}(d,g)\ge 0.5,\\
\mathrm{known}, &
\max_{g\in G_k}\mathrm{IoU}(d,g)\ge 0.5,\\
\mathrm{neg}, &
\max_{g\in G_k\cup G_u}\mathrm{IoU}(d,g)<0.3,\\
\mathrm{amb}, &
\text{otherwise}.
\end{cases}
\]
Thus, proposals matching future-task objects are treated as positive
unknowns; proposals matching current-task objects are known-as-unknown
errors; proposals far from all ground truth are background-type
negative unknowns; and the remaining intermediate-overlap cases are
ambiguous.

Let $\mathcal{D}$ be the raw unknown predictions, $\mathcal{K}\subseteq
\mathcal{D}$ the predictions retained after filtering, and
$\mathcal{I}$ the evaluation images. We report
\[
\mathrm{FUPI}
=
\frac{
|\{d\in\mathcal{K}:\ell(d)=\mathrm{neg}\}|
}
{|\mathcal{I}|},
\]
which measures the number of retained background-type false unknowns
per image. FUPI does not include known-as-unknown predictions.
The suppression gain is
\[
\mathrm{SG}
=
1-
\frac{
|\{d\in\mathcal{K}:\ell(d)=\mathrm{neg}\}|
}
{
|\{d\in\mathcal{D}:\ell(d)=\mathrm{neg}\}|
}.
\]
Novel mis-suppression harm is proposal-level:
\[
\mathrm{NMH}
=
\frac{
|\{d\in\mathcal{D}\setminus\mathcal{K}:\ell(d)=\mathrm{pos}\}|
}
{
|\{d\in\mathcal{D}:\ell(d)=\mathrm{pos}\}|
}.
\]
For standard OWOD metrics, we follow prior definitions: U-Recall is computed
as in OWOD~\cite{vit}, and UDP follows RE-OWOD~\cite{re_owod}. For raw
detectors we evaluate these metrics on the original unknown stream
$\mathcal{D}$; for post-hoc filters we evaluate them on the retained stream
$\mathcal{K}$ after suppression. Thus, FUPI and NMH are newly introduced
proposal-level diagnostics, whereas U-Recall and UDP are inherited metrics
reported under the same retained-output protocol.

\subsection{Main Results}
\label{sec:main_results}

\begin{figure}[t]
    \centering
    \includegraphics[width=0.75\linewidth,height=0.42\textheight,keepaspectratio]{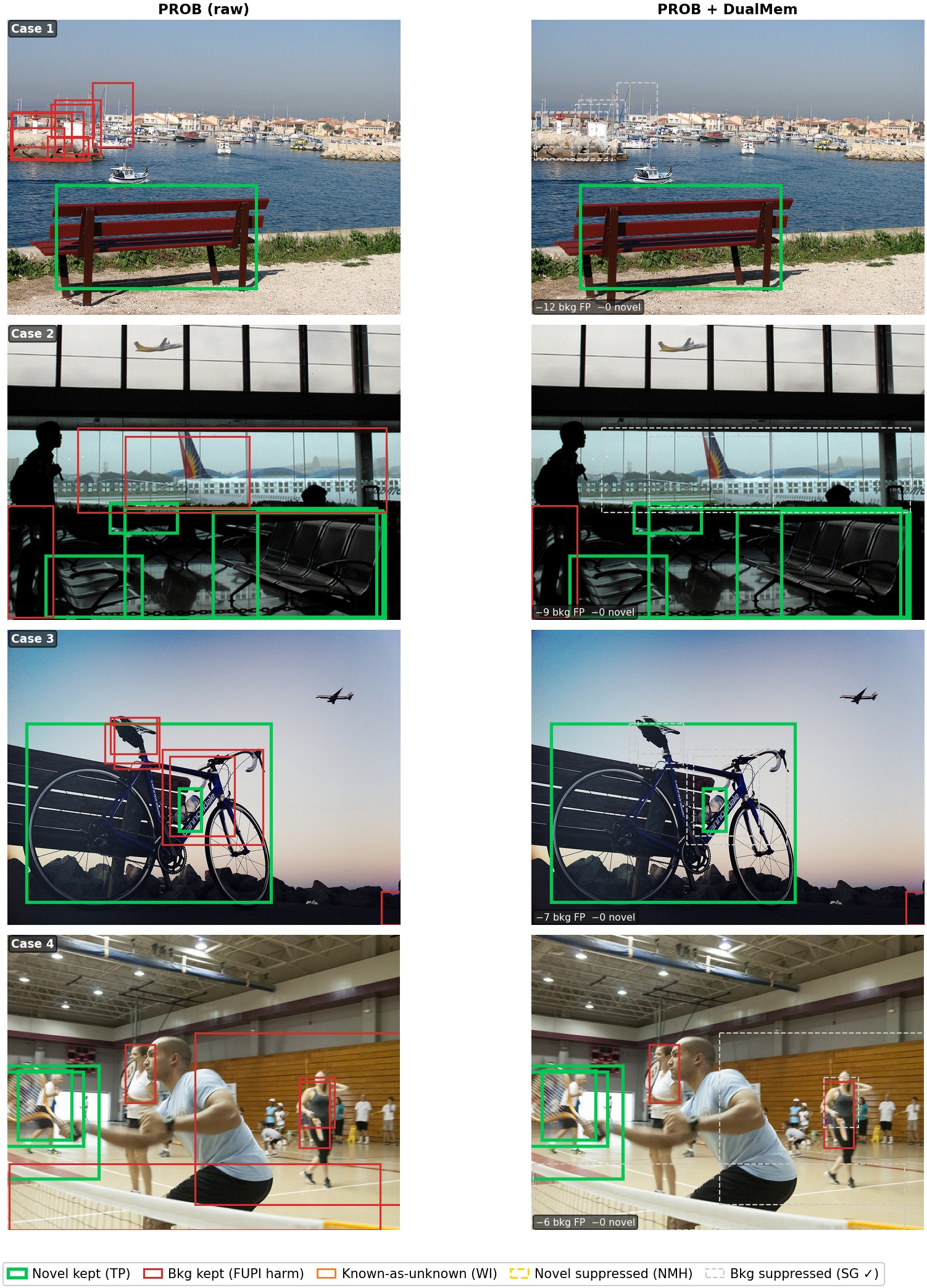}
    \caption{
    Qualitative success cases on M-OWODB Task~1. Each row pairs raw
    PROB output (left) with PROB$+$\methodname{} (right). Green boxes
    mark retained novel proposals, red boxes retained background false
    positives, and dashed white boxes background proposals suppressed by
    \methodname{}. Across these scenes, \methodname{} removes $6$--$12$
    background false positives per image while suppressing zero positive
    unknowns.
    }
    \label{fig:success_cases}
\end{figure}

Table~\ref{tab:main_results} reports \methodname{} across three OWOD
detectors on M-OWODB Task~1. \methodname{} reduces FUPI substantially on
all three detectors: by $58.6\%$ on PROB ($7.35 \to 3.04$), $44.9\%$ on
OW-DETR ($2.83 \to 1.56$), and $66.3\%$ on HypOW ($5.23 \to 1.76$). UDP
improves correspondingly, from $0.114 \to 0.258$ on PROB, $0.023 \to
0.042$ on OW-DETR, and $0.123 \to 0.270$ on HypOW. The mean FUPI reduction across the three detectors is $56.6\%$. On PROB
Task~1, \methodname{} achieves a $58.6\%$ FUPI reduction, more than doubling
the $24.5\%$ reduction achieved by the K-means prototype baseline
(Table~\ref{tab:dual_memory_ablation}).

NMH remains close to the calibration target $\alpha=0.10$ on PROB
($6.4\%$) and OW-DETR ($7.5\%$). On HypOW, actual NMH ($22.6\%$) exceeds
the target. We attribute this gap to a distribution mismatch between
HypOW's positive unknowns---which concentrate near low-density regions in
SigLIP space---and the calibration positives used for quantile selection.
The NP framework makes this case explicit and addressable: using a more conservative $\alpha$ on HypOW can reduce the realized NMH, but
requires per-detector validation. Crucially, because \methodname{} is applied only to predictions assigned to the unknown stream and known-class predictions bypass the filter, known-class
mAP is unchanged under the standard evaluation pipeline. Figure~\ref{fig:success_cases} shows four representative scenes in which \methodname{} removes $6$--$12$ background false positives per image while preserving all positive unknowns---a visual counterpart to the FUPI and NMH numbers in Table~\ref{tab:main_results}.

\begin{table}[t]
    \centering
    \small
    \setlength{\tabcolsep}{4pt}
        \resizebox{\linewidth}{!}{%
    \begin{tabular}{@{}llrrrrr@{}}
        \toprule
        Method & Detector & FUPI$\downarrow$ & UDP$\uparrow$ & NMH & U-Recall & $\Delta$U-Rec.\\
        \midrule
        Raw            & PROB    & 7.35 & 0.114 & --     & 0.218 & --   \\
        + \methodname{} & PROB    & \textbf{3.04} & \textbf{0.258} & 6.4\%  & 0.206 & --1.2 \\
        \midrule
        Raw            & OW-DETR & 2.83 & 0.023 & --     & 0.020 & --   \\
        + \methodname{} & OW-DETR & \textbf{1.56} & \textbf{0.042} & 7.5\%  & 0.018 & --0.2 \\
        \midrule
        Raw            & HypOW   & 5.23 & 0.123 & --     & 0.179 & --   \\
        + \methodname{} & HypOW   & \textbf{1.76} & \textbf{0.270} & 22.6\% & 0.145 & --3.4 \\
        \midrule
        \textbf{Mean reduction} & -- & \textbf{$-56.6\%$} & \textbf{$+119.2\%$} & -- & -- & $-1.6$\,pp \\
        \bottomrule
    \end{tabular}%
    }
\caption{
    Main results on M-OWODB Task~1. \methodname{} substantially reduces background-type   false-unknown density and improves UDP across three detectors at
    $\alpha=0.10$. NMH stays close to the calibration target on PROB and
    OW-DETR; on HypOW it exceeds the target, indicating a calibration--test mismatch and
the need for per-detector validation.
    }
    \label{tab:main_results}
\end{table}

\subsection{Why Dual Memory? Ablation}
\label{sec:dual_memory_ablation}

Table~\ref{tab:dual_memory_ablation} compares \methodname{} with two
baselines on PROB Task~1: a tuned objectness threshold, and a
\emph{KMeans prototype baseline} that compresses each memory into
$K$ centroids and uses a max-cosine dual-threshold rule
($\mathcal{M}^+$: $K{=}16$; $\mathcal{M}^-$: $K{=}64$; $\tau{=}0.80$).
This baseline isolates the contribution of our three design
choices---$k$-NN reference set, LRT smoothing, NP
calibration---against a natural prototype-compression alternative.
The upgraded \methodname{} achieves a $45.2\%$ lower FUPI than
the prototype baseline ($5.55 \to 3.04$) while \emph{simultaneously}
reducing NMH ($7.4\% \to 6.4\%$). Compared with objectness thresholding,
\methodname{} reaches a much lower FUPI ($3.04$ vs.\ $5.85$) at less than
half the NMH ($6.4\%$ vs.\ $13.7\%$). The improvement decomposes as follows: replacing KMeans centroids with the full $k$-NN
reference set is the dominant gain (FUPI $5.55 \to 3.12$ at $T\to 0$,
$-43.6\%$); LRT temperature smoothing with $T=0.05$ provides
additional refinement ($3.12 \to 3.04$, $-2.6\%$ relative); and NP
calibration replaces a magic threshold with an explicit, monotonic
$\alpha$-controlled trade-off.

\begin{table}[t]
    \centering
    \small
    \setlength{\tabcolsep}{3pt}
        \resizebox{0.98\linewidth}{!}{%
    \begin{tabular}{@{}lrrrr@{}}
        \toprule
        Strategy & FUPI$\downarrow$ & NMH$\downarrow$ & U-Recall & UDP$\uparrow$ \\
        \midrule
        Raw PROB                                  & 7.35 & --     & 0.218 & 0.114 \\
        + Objectness threshold ($\geq 0.60$)      & 5.85 & 13.7\% & 0.204 & 0.139 \\
        + KMeans prototype baseline ($\tau{=}0.80$) & 5.55 & 7.4\%  & 0.207 & 0.150 \\
        + \methodname{} (LRT$+k$NN$+$NP, $\alpha{=}0.10$) & \textbf{3.04} & \textbf{6.4\%} & 0.206 & \textbf{0.258} \\
        \bottomrule
    \end{tabular}%
    }
\caption{
Decision-rule ablation on PROB Task~1. The upgraded \methodname{}
improves on both the KMeans prototype baseline and a tuned
objectness threshold in FUPI and NMH simultaneously.
}
    \label{tab:dual_memory_ablation}
\end{table}

\subsection{Critic Choice}
\label{sec:critic_ablation}

Table~\ref{tab:critic_ablation} compares frozen critics on PROB
Task~1 under the upgraded framework. The most striking observation
concerns the detector's own decoder query: \emph{despite the highest
linear-probe AUROC of all tested features} ($0.908$;
Table~\ref{tab:cross_detector_probe}), it is the \emph{worst} single critic
under memory-based retrieval (FUPI $3.40$, NMH $10.2\%$). DINOv2
shows the same pattern in a milder form: a slightly higher
linear-probe AUROC than SigLIP ($0.879$ vs.\ $0.871$, not shown in the table),
but a higher
NMH ($8.2\%$ vs.\ $6.4\%$) under cosine $k$-NN retrieval~\cite{dinov2}.
This is a
direct empirical decoupling: AUROC measures separability \emph{after}
an optimal supervised one-dimensional projection, while
\methodname{} relies on the \emph{raw} geometry of the feature
space. PROB's decoder query encodes detection-specific spatial and
class information shaped by the DETR training objective~\cite{detr}; its raw
cosine structure carries discriminative signal only after a learned
projection. By contrast, CLIP/SigLIP-style language--image pretraining
aligns semantic similarity with feature-space geometry~\cite{clip,siglip},
yielding a feature space in which
$k$-NN retrieval directly captures object identity.

Among the single critics, SigLIP achieves the best NMH ($6.4\%$,
closest to the calibration target $\alpha=0.10$) and the highest
U-Recall ($0.206$), at a modest FUPI cost ($3.04$ vs.\ $2.88$ for
DINOv2). We adopt SigLIP as the default. Fused critics
(SigLIP$\Vert$PROB concat, SigLIP$+$PROB average) reduce FUPI
further to $2.35$ but at the cost of higher dimensionality
($1024$-D / averaged $256$-D) and higher NMH; we report these for
completeness but do not use them as the default. CLIP ViT-B/32~\cite{clip} yields
the lowest FUPI ($1.53$) but with $24.8\%$ NMH, suppressing one in
four positive unknowns and dropping U-Recall by $5$pp---a clear case
where aggressive false-positive cleanup overrides the NP budget,
and a useful negative example of how a misaligned critic geometry
can subvert the calibration framework.

\paragraph{Implication.} The decoupling between linear-probe AUROC
and memory-based filtering effectiveness has a practical
consequence: \emph{the natural-looking idea of using the detector's
internal decoder query as a critic does not work}, even though the
information is provably there ($\mathrm{AUROC}=0.908$). Recovering
this information requires a feature space whose raw geometry is
aligned with semantic similarity---which is precisely what a
contrastively pretrained external critic provides, and what the
detector's training objective does not. This complements our main
information-bottleneck thesis: the loss at the objectness head is
not the only structural problem; the decoder query itself, while
information-rich, is geometrically unsuited for the post-hoc
critic role.

\begin{table}[t]
    \centering
    \small
    \setlength{\tabcolsep}{4pt}
        \resizebox{\linewidth}{!}{%
    \begin{tabular}{@{}lrrrrr@{}}
        \toprule
        Critic feature        & Dim. & FUPI$\downarrow$ & UDP$\uparrow$ & NMH$\downarrow$ & U-Recall \\
        \midrule
        Raw PROB              & --   & 7.35 & 0.114 & --     & 0.218 \\
        \midrule
        PROB decoder query    & 256  & 3.40 & 0.204 & 10.2\% & 0.206 \\
        DINOv2 ViT-B/14       & 768  & 2.88 & 0.268 & 8.2\%  & 0.205 \\
        \textbf{SigLIP ViT-B/16}  & 768 & \textbf{3.04} & 0.258 & \textbf{6.4\%} & \textbf{0.206} \\
        \midrule
        SigLIP $+$ PROB (avg) & 256  & 2.35 & 0.303 & 9.2\%  & 0.205 \\
        SigLIP $\Vert$ PROB (concat) & 1024 & 2.35 & 0.305 & 7.1\% & 0.206 \\
        \midrule
        CLIP ViT-B/32         & 512  & 1.53 & 0.303 & \textcolor{red}{24.8\%} & \textcolor{red}{0.168} \\
        \bottomrule
    \end{tabular}%
    }
\caption{
    Critic ablation on PROB Task~1 under the LRT$+k$NN$+$NP framework. 
    SigLIP achieves the best NMH and the highest U-Recall, and is
    used as the default critic. CLIP yields lower FUPI but
    higher NMH ($24.8\%$); we report it as a stress test, not
    as a recommendation.
    }
    \label{tab:critic_ablation}
\end{table}

\subsection{User-Controllable Trade-off: $\alpha$ Sweep}
\label{sec:alpha_sweep}

\begin{figure}[!htbp]
    \centering
    \includegraphics[width=0.95\linewidth]{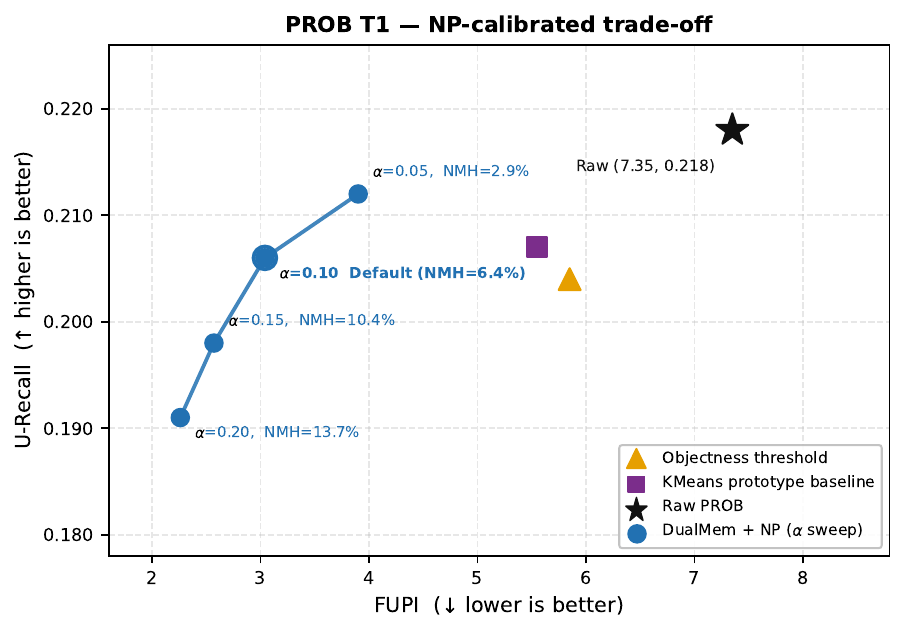}
    \caption{
    NP-calibrated $\alpha$ sweep on PROB Task~1. Each operating point
    corresponds to a user-specified false-suppression budget $\alpha$.
    The curve monotonically traces a precision--recall trade-off; the
    raw detector and a tuned objectness threshold are shown for
    reference. The default $\alpha=0.10$ is marked.
    }
    \label{fig:alpha_sweep}
\end{figure}

A key advantage of the NP framework is that $\tau$ is no longer a magic
constant. Figure~\ref{fig:alpha_sweep} shows the $\alpha$ sweep on PROB
Task~1: $\alpha$ monotonically trades FUPI for U-Recall, and actual NMH
on the test split is consistently \emph{below} the target budget on
PROB, reflecting a conservative (user-favorable) calibration gap. Users
can pick an operating point matching their downstream tolerance for
novel mis-suppression---e.g., $\alpha=0.05$ yields NMH $=2.9\%$ at the
cost of higher FUPI ($3.90$), while $\alpha=0.20$ pushes FUPI down to
$2.26$ at NMH $=13.7\%$.

\section{Discussion and Limitations}
\label{sec:discussion}

\methodname{} is a post-hoc filter, not a replacement for training-time
OWOD objectives. Some methods directly suppress false unknowns during
training (e.g., UnSniffer~\cite{unsniffer}); applying \methodname{} on
top of such a detector yields near-zero additional gain, indicating
that the two regimes target overlapping failure modes. \methodname{} is
most useful when the detector is fixed: proprietary checkpoints, large
pretrained OWOD models, or legacy detectors. In this setting, the
frozen external critic recovers information that the detector's
one-dimensional unknown confidence does not expose.

\paragraph{Detector-internal residual.} A natural extension is to add a
trained residual head $g_\theta(q(p))$ on top of $\Lambda(p)$, where
$q(p)$ is the detector's 256-D decoder query. On PROB Task~1, this
yields negligible additional gain (FUPI $3.04 \to 3.03$, $<1\%$). We
interpret this---together with its lack of cross-detector
applicability---as further evidence for the information-bottleneck
thesis: the detector-internal signal and the SigLIP signal recover
overlapping discriminative content, and stacking them does not produce
additive gains. We therefore exclude the residual head from the main
\methodname{} configuration.

\begin{figure}[t]
    \centering
    \includegraphics[width=\columnwidth]{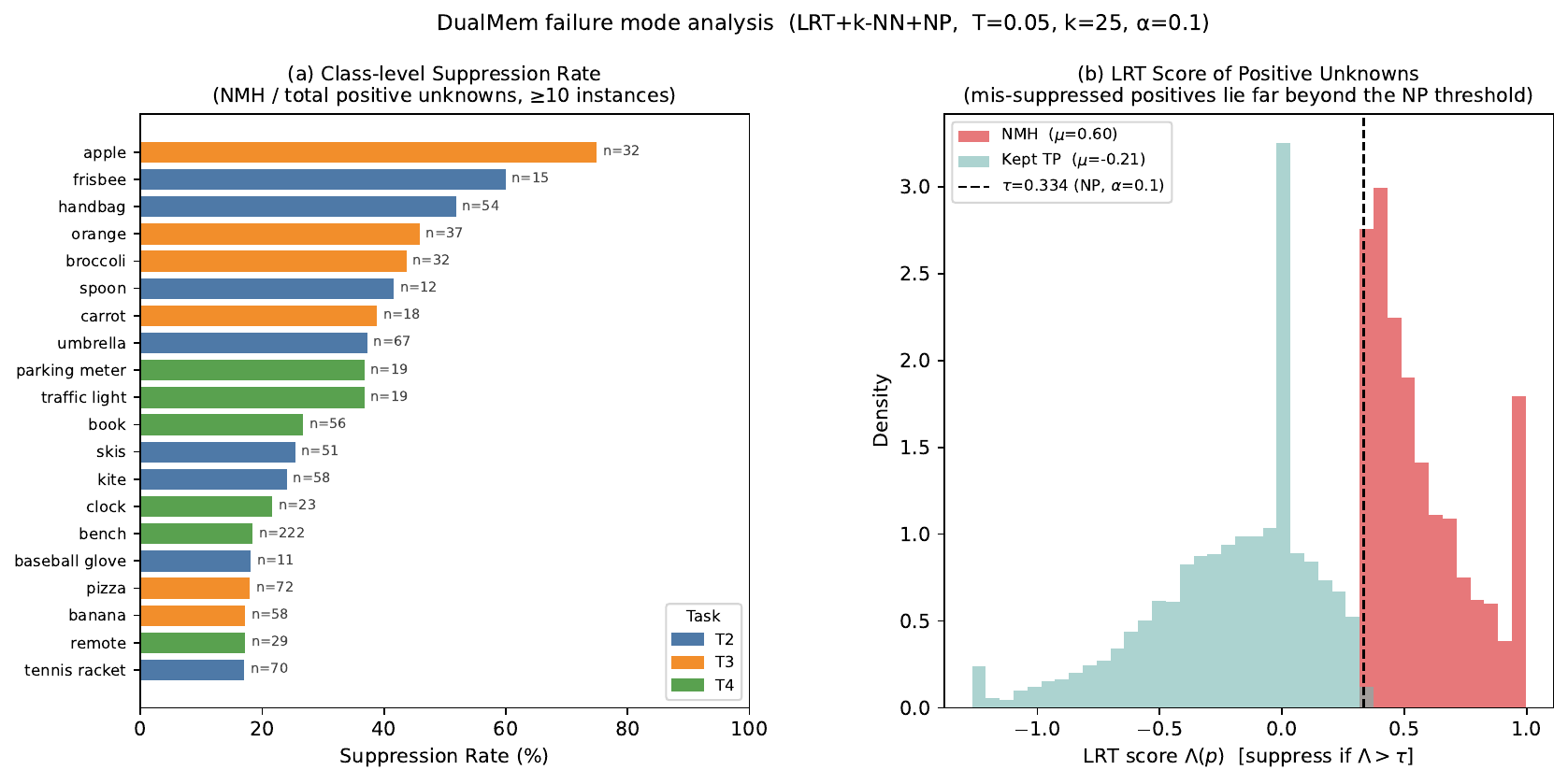}
    \caption{
    Failure analysis of DualMem on positive unknown proposals under the default LRT+kNN+NP configuration
    ($T=0.05$, $k=25$, $\alpha=0.10$).
    (a) Mis-suppression is class-concentrated rather than uniform; only classes with at least ten positive unknown proposals are shown; rates for low-count classes should be interpreted as descriptive rather than 
statistically stable.
    (b) Mis-suppressed positives have substantially higher LRT scores than retained true-positive unknowns and lie beyond the NP-calibrated threshold, indicating that most failures are not marginal boundary cases.
    }
    \label{fig:failure_analysis}
\end{figure}

\paragraph{Failure analysis.}
Figure~\ref{fig:failure_analysis} analyzes positive unknown proposals suppressed by DualMem on PROB. 
Mis-suppression is not uniformly distributed across classes: several visually context-dependent categories, such as apple, frisbee, handbag, orange, and broccoli, exhibit substantially higher suppression rates. 
In the decision space, mis-suppressed positives are shifted well beyond the NP threshold, whereas retained positives concentrate mostly below it. 
This suggests that DualMem failures are not merely boundary cases, but arise when true novel objects are represented as background-like under the frozen critic geometry.

\paragraph{Limitations.} We note three. (i) The NP budget does not transfer uniformly across detectors: PROB and OW-DETR stay near the target, whereas HypOW exhibits a larger calibration--test gap. (ii) As shown in Fig.~\ref{fig:failure_analysis}, suppressed positives are concentrated in a small set of visually context-dependent categories, and their LRT scores lie well beyond the NP threshold. This points to a limitation of the frozen critic geometry rather than objectness alone.
 (iii) DualMem relies on a small annotated calibration split to build
the positive memory $M^+$. This is weaker than retraining the detector
on future classes, since detector parameters remain frozen, but it is
not a fully annotation-free OWOD setting. The method is therefore most
appropriate for calibrated post-hoc deployment, auditing, or model
maintenance scenarios where a small set of novel-object annotations is
available.

\section{Conclusion}
\label{sec:conclusion}

OWOD detectors expose a weak scalar interface for unknown detection:
across PROB, OW-DETR, and HypOW, the final objectness score is a poor
positive-vs-negative unknown discriminator, whereas detector-internal
proposal features and frozen SigLIP crop features retain substantially
stronger separability. This identifies a cross-detector objectness
bottleneck: discriminative proposal-level information exists, but is
not reliably exposed by the final one-dimensional unknown confidence.

DualMem addresses this bottleneck as a calibrated post-hoc filter. It
performs a non-parametric likelihood-ratio test in frozen SigLIP
feature space using positive and negative k-nearest-neighbor memories,
and selects the suppression threshold through Neyman--Pearson
calibration. Across PROB, OW-DETR, and HypOW on M-OWODB Task 1, DualMem
reduces background false unknown density by 44.9--66.3\% (mean 56.6\%)
while keeping U-Recall within 3.4 percentage points; known-class mAP is
unchanged because known detections bypass the filter. These results suggest that
post-hoc unknown-stream cleanup is a useful complement to training-time
OWOD objectives, especially when the detector is fixed.

\bibliography{aaai2027}

\end{document}